\documentclass[runningheads]{llncs}
\usepackage{graphicx}
\usepackage{tikz}
\usepackage{comment}
\usepackage{amsmath,amssymb} % define this before the line numbering.
\usepackage{color}
\usepackage{multirow}
\usepackage[accsupp]{axessibility}  % Improves PDF readability for those with disabilities.
\usepackage{subfigure}
\usepackage{orcidlink}

\begin{document}

\newcommand{\para}[1]{\vspace{1.5mm}\noindent\textbf{#1}}

% % Customized marcos for review.
% \newcommand{\ca}[1]{\textcolor[rgb]{1.0,0.0,0.0}{#1}}
% \newcommand{\md}[1]{\textcolor[rgb]{0.0,0.0,1.0}{#1}}
% \newcommand{\shc}[1]{\textcolor[rgb]{0.0,0.8,0.0}{(Sheng: #1)}}

% \renewcommand\thelinenumber{\color[rgb]{0.2,0.5,0.8}\normalfont\sffamily\scriptsize\arabic{linenumber}\color[rgb]{0,0,0}}
% \renewcommand\makeLineNumber {\hss\thelinenumber\ \hspace{6mm} \rlap{\hskip\textwidth\ \hspace{6.5mm}\thelinenumber}}
% \linenumbers
\pagestyle{headings}
\mainmatter
\def\ECCVSubNumber{2025}  % Insert your submission number here

\title{SuperLine3D: Self-supervised Line Segmentation and Description for LiDAR Point Cloud
} % Replace with your title
% INITIAL SUBMISSION 
\begin{comment}
\titlerunning{ECCV-22 submission ID \ECCVSubNumber} 
\authorrunning{ECCV-22 submission ID \ECCVSubNumber} 
\author{Anonymous ECCV submission}
\institute{Paper ID \ECCVSubNumber}
\end{comment}
%******************

% CAMERA READY SUBMISSION
% \begin{comment}
\titlerunning{SuperLine3D: Self-supervised Line Segmentation and Description}
% If the paper title is too long for the running head, you can set
% an abbreviated paper title here
%
\author{Xiangrui Zhao\inst{1,2}\orcidlink{0000-0002-0129-1933}\and
Sheng Yang\inst{2}\and
Tianxin Huang\inst{1}\and
Jun Chen\inst{1}\and \\
Teng Ma\inst{2}\and
Mingyang Li\inst{2}\and
Yong Liu\inst{1, \dag}\orcidlink{0000-0003-4822-8939}}
\authorrunning{X. Zhao et al.}
\institute{APRIL Lab, Zhejiang University, China\\
\email{\{xiangruizhao, 21725129, junc\}@zju.edu.cn, yongliu@iipc.zju.edu.cn} \and
Autonomous Driving Lab, DAMO Academy, China\\
\email{\{shengyang93fs, mingyangli009\}@gmail.com, damon.mt@alibaba-inc.com}}

% \end{comment}
%******************
\maketitle

\makeatletter
\def\blfootnote{\xdef\@thefnmark{}\@footnotetext}
\makeatother

\blfootnote{$^\dag$ indicates the corresponding author.}

\begin{abstract}

% % What we do.
% We propose the first learning-based line feature detector and descriptor for LiDAR point clouds.
% Motivation and Method.
Poles and building edges are frequently observable objects on urban roads, conveying reliable hints for 
%vehicle localization. 
{various computer vision tasks.} 
To repetitively extract them as features 
%to 
{and perform association}
between discrete LiDAR frames for registration, we propose the first learning-based feature segmentation and description model for 3D lines in LiDAR point cloud.
% and then associate through discrete LiDAR frames for SLAM, we propose two models for line feature segmentation and description, respectively.
%%%! Detector现在只讲了做了什么事情。我们需要在摘要中说明Detector这个阶段的创新点：是训练过程参照了SuperPoint放在这里，还是什么别的Fancy Method/Idea可以放在这里？
To train our model without {the time consuming and tedious data labeling process,}
we first generate synthetic primitives for the basic appearance of target lines, and build {an iterative line auto-labeling process to gradually refine} line labels on real LiDAR scans. Our segmentation model can extract lines under arbitrary scale perturbations, and we use {shared EdgeConv encoder layers to train the two segmentation and descriptor heads jointly}.
Base on the model, we can build a highly-available global registration module for point cloud registration, in conditions without initial transformation hints. 
Experiments have demonstrated that our line-based registration method is highly competitive to state-of-the-art point-based approaches. Our code is available at \url{https://github.com/zxrzju/SuperLine3D.git}.
% 说不用RANSAC的点好，太过于细节了，可以在Experiments中说，毕竟RANSAC是它们的一部分。
    \keywords{3D Line Feature, Point Cloud Registration}
\end{abstract}
\section{Introduction} \label{sec:intro}

% Point cloud registration.
Point cloud registration is an essential technique for LiDAR-based vehicle localization on urban road scenes~\cite{pomerleau2015review}. Considering recent researches~\cite{huang2021comprehensive,gu2020review}, the SLAM community~\cite{khan2021comparative} divides these algorithms into two categories regarding their purpose, as \emph{local} and \emph{global} search methods, respectively.
% The local search class.
The \emph{local} search category~\cite{besl1992method,biber2003normal} typically constructs a non-convex optimization problem by greedily associating nearest entities to align. This often relies on a good initial guess, and thus mostly used for incremental positioning modules such as the LiDAR odometry~\cite{zhang2014loam} and map-based localization~\cite{rozenberszki2020lol}.
% The global search class.
The \emph{global} search category {is used for less informative conditions}, i.e., relocalization and map initialization problems when the initial guess is not reliable and large positional and rotational change exists. Since nearest neighbor search methods cannot find correct matching pairs in the Euclidean space, \emph{global} search algorithms choose to extract distinct entities and construct feature descriptors~\cite{zhou2016fast}, to establish matches in the description space.

% Global search, feature detection and description.
There exists a variety of classical hand-crafted features (e.g., FPFH~\cite{rusu2009fast}) for global search and registration, and recent learning-based methods~\cite{zhang2020deep} have improved {the registration accuracy and success rate}. However, the performance of some methods~\cite{pais20203dregnet,perez2019deep} severely drops when adapting to {real LiDAR scans, because the density of scanned points is inversely proportional to the scanning distance, and thus influences the coherence of point description.} Considering such limitation of a single point, we propose an idea of using structural {lines}, analogously as previous approaches proposed for images~\cite{huang2020tp,zhang2021elsd}, to see {whether a relatively stable descriptor can be concluded through a semantically meaningful group of scattered points.}

% Objectives.
In typical LiDAR point cloud scanned from urban road scenes, there are three categories of lines. 1) Intersection of planes, e.g., edge of two building facades and curbs. 2) Standalone pole objects, e.g., street lamps and road signs alongside the road. 3) Virtual line {consists of edge points across multiple scan rings, generated by ray-obstacle occlusions.} While the last category is not repeatable and thus inappropriate for localization, the first two types are practical landmarks suitable to be extracted and described. Since these line segments are {larger} targets compared to point features, they have a higher chance to be {repeatably observed}. Moreover, {the concluded position of each line is more precise to a single corresponding point between frames due to the limited scanning resolution, which causes sampling issues.}

\begin{figure}[ht]
    \centering
    \subfigure[Automatic Line Segments Labeling]{
        \includegraphics[height=3.55cm]{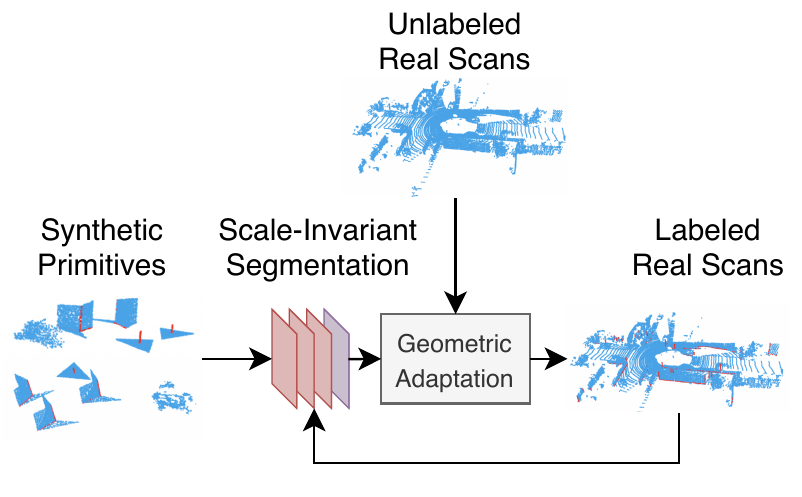}
    }
   \subfigure[Line Segmentation and Description]{
        \includegraphics[height=3.55cm]{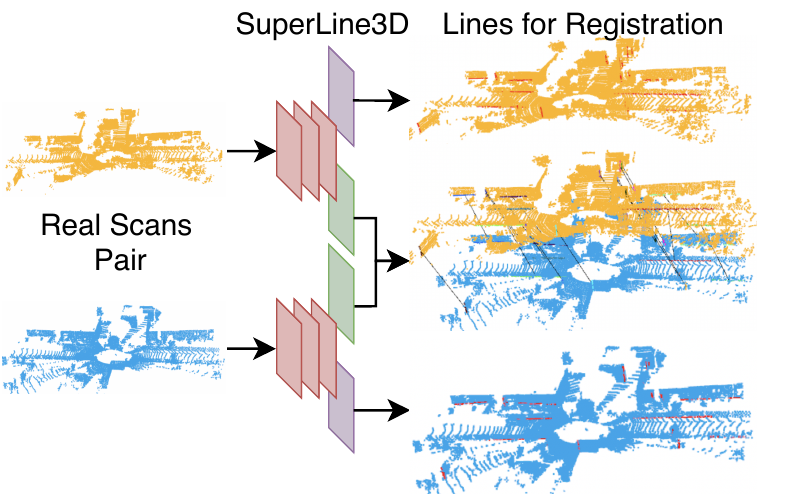}
    } 
    \caption{Pipeline overview. a): We train a scale-invariant segmentation on the synthetic data and get the precise line segment labels after multiple geometric adaptation iterations. b): We simultaneously train segmentation and description on labeled LiDAR scans, where {red, purple, and green layers stand for encoders, segmentation header, and description header, respectively}.}
    \label{fig:pipeline}
\end{figure}

% Contribution.
In this paper, we propose a self-supervised learning method for line segmentation and description on LiDAR scans (Fig.~\ref{fig:pipeline}). 
% \ca{For extracting lines on frames, we train a scale-invariant segmentation model on labeled synthetic data to automatically label the LiDAR scans. Then, for segmenting and describing lines between scans, we jointly train line segmentation and description on the labeled LiDAR scans.} 
Following the training procedure of SuperPoint~\cite{detone2018superpoint} to solve the lack of publicly available training data, we choose to train our line extraction model, by first construct limited synthetic data and then perform {auto labeling} on real scans. {By sharing point cloud encoding layers and use two separate branches for decoding and application headers, we are able to jointly train two tasks on those generated data.} We view such a pipeline to train and use line features for the scan registration purpose as the key contribution of our work, which includes:

\begin{itemize}
    \item From the best of our knowledge, we propose the first learning-based line segmentation and description for LiDAR {scans, bringing up an applicable} feature category for global registration.
    \item We propose a line segment labeling method for point clouds, which can migrate the model learned from synthetic data to real LiDAR scans for automatic labeling.
    \item We explore the scale invariance of point cloud features, and provide a feasible idea for improving the generalization of learning-based tasks on the point cloud under scale perturbations by eliminating the scale factor in Sim(3) transformation.%, \ca{how? use one sentence}. % rather than only relying on data augmentation.
    % \item Extensive experimental results show that our line features can maintain high \md{repeatability and success rate of association} under large-angle perturbations, which also generalize well on unseen datasets.
\end{itemize}

Extensive experimental results {have shown} that our line-based registration can maintain high success rate and accuracy under large-angle perturbations, and the trained model on one real scans dataset is highly adaptable to other {urban scene} datasets.

\section{Related Work} \label{Related}

\para{Learning-based Point Cloud Registration.} In recent researches, there are a variety of learning-based approaches proposed for registering point clouds, and we can divide them into two groups considering whether explicit features have been extracted. End-to-end approaches use ground-truth transformation in loss calculation, and predict the transformation directly through the network: FMR~\cite{huang2020feature} registers point clouds by minimizing feature-metric loss, and PCRNet~\cite{sarode2019pcrnet} evaluate the similarity of PointNet~\cite{qi2017pointnet} features and regresses poses through fully connected layers directly. {These trained end-to-end models work well on tested sequences, but they are facing a practical problem on how to perform a joint state estimation in a multi-sensor fusion system~\cite{fang2005multi}. Nevertheless, knowledge of these models are hardly adaptable to different motion scheme and other datasets. Therefore, methods with explicit feature extraction and description are still an active branch in the SLAM community.}

\para{Registration with Explicit Features.} Start with hand-crafted features (e.g., FPFH~\cite{rusu2009fast} and ISS~\cite{zhong2009intrinsic}) concluding local patch appearances of point clouds, methods of extracting and describing explicit features mainly aim at the saliency of entities and coherency of description. While hand-crafted features are mostly designed for evenly sampled clouds, learning-based features~\cite{li2019usip,choy2019fully,pais20203dregnet,choy2020deep,lu2021hregnet,bai2021pointdsc} have better robustness and generalization, once trained on the target LiDAR scan datasets.
D3Feat~\cite{bai2020d3feat} uses kernel-based convolution~\cite{thomas2019kpconv} to learn feature detection and description. SpinNet~\cite{ao2021spinnet} builds a rotation-invariant local surface descriptor through novel spatial point transformer and 3D cylindrical convolutional layer. Both D3Feat~\cite{bai2020d3feat} and SpinNet~\cite{ao2021spinnet} are state-of-the-art learning-based point features, but they still suffer from the inherent problem of point features, {and thus requires sample consensus as a post pruning procedure} to filter correct feature associations.

\para{Line Features for SLAM.} Image based line-aware approaches for detection (e.g., LSD~\cite{von2012lsd}, EDLines~\cite{akinlar2011edlines}, and TP-LSD~\cite{huang2020tp}), description (e.g., LBD~\cite{zhang2013efficient} and SOLD2~\cite{pautrat2021sold2}), and systematical SLAM designs (e.g., PL-SLAM~\cite{pumarola2017pl}) have been well studied in recent years, whereas LiDAR scan based extraction and description methods, although heavily used in modern LiDAR SLAM approaches (e.g., LOAM~\cite{zhang2014loam} and LeGO-LOAM~\cite{shan2018lego}), are under explored. To the best of our knowledge, we found Lu et al.~\cite{lu2019fast} have proposed a 3D line detection method through projecting LiDAR points onto image, and thus convert the task into a 2D detection problem. Chen et al.~\cite{chen2021efficient} based on this work~\cite{lu2019fast} to carry out a line-based registration approach for structural scenes.
However, their limitations are two folds: 1) only work on organized point clouds, and 2) have not addressed line description and thus not suitable for global search registration problems. In contrast, we follow the idea of descriptor conclusion from SOLD2~\cite{pautrat2021sold2}, which has been proven to be useful in our paper for the coherency of describing a group of points.

\section{Method} \label{Method}

Considering the lack of available labeled line datasets of LiDAR scans, we follow the self-supervised idea of SuperPoint~\cite{detone2018superpoint}, to train our line segmentation model, by first constructing a simple synthetic data to initialize a base model, and then {refining} the model iteratively with {auto-labeled} real LiDAR scans {from} geometric adaptation (Sec.~\ref{method:line_label}).
%These line labels are clusters of input points stand for multiple extracted line segments.
After that, we gather line correspondences between different LiDAR scans, and jointly train the line segmentation and description in an end-to-end approach (Sec.~\ref{method:desc}).
% \ca{, and thus we perform self-supervised line segment labeling for generating the ground-truth of frame-wise associations}. 
% After that, we gather these inferenced line detection results to learn their corresponding descriptor. The whole two-stage models are trained in an end-to-end approach (Sec.~\ref{method:desc}).

% There are many strong baselines~\cite{huang2020tp,zhang2021elsd} and datasets~\cite{huang2018learning,denis2008efficient} for line features in images, while so far, there is no publicly-available dataset for line features in the point cloud. Inspired by $SuperPoint$~\cite{detone2018superpoint} and $SOLD2$~\cite{pautrat2021sold2}, we construct a self-supervised line segment detection and description pipeline for the point cloud. First, we train a base detector on labeled synthetic data. Then the trained model is used to label the LiDAR point cloud automatically. Finally, we train the line detection and description in an end-to-end approach.

\subsection{Line Segmentation Model}~\label{method:det}

\begin{figure}
    \centering
    \subfigure[Mesh Model]{
        \includegraphics[width=0.19\textwidth]{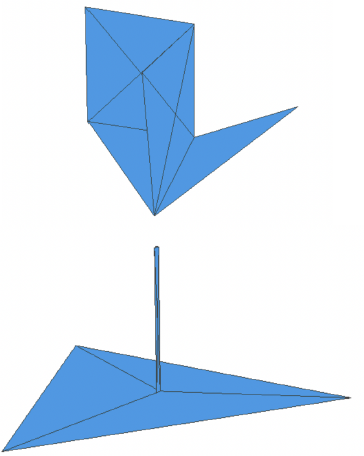}
        \label{fig:synthetic_data:mesh}
        }
    \subfigure[Point Cloud]{
        \includegraphics[width=0.19\textwidth]{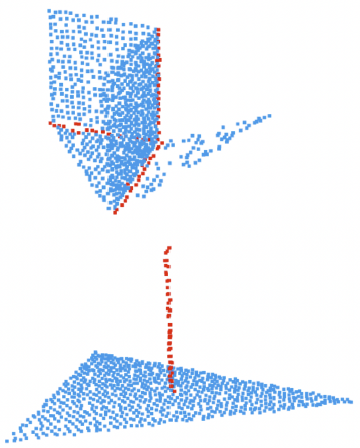}
        \label{fig:synthetic_data:pc}
        }
    \subfigure{
        \begin{minipage}[b]{0.25\textwidth}
            \setcounter{subfigure}{2}
            \subfigure[Noise Sampling]{
            \includegraphics[width=\textwidth]{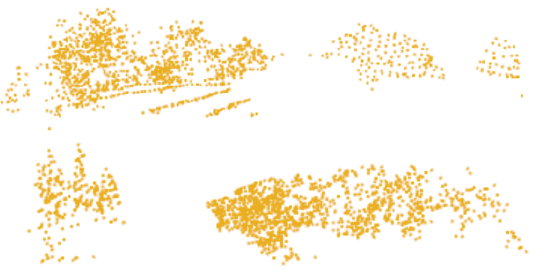}
            }
            \subfigure[Synthetic Data]{
            \includegraphics[width=\textwidth]{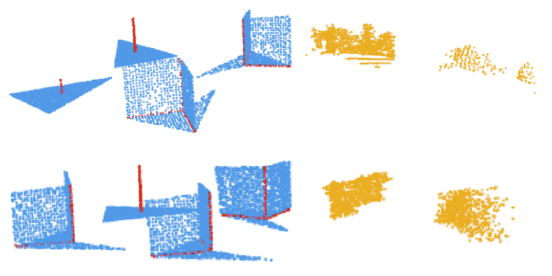}
            }
        \end{minipage}
        \label{fig:synthetic_data:noise}
    }
    \subfigure[Scale Comparision]{
        \includegraphics[width=0.25\textwidth]{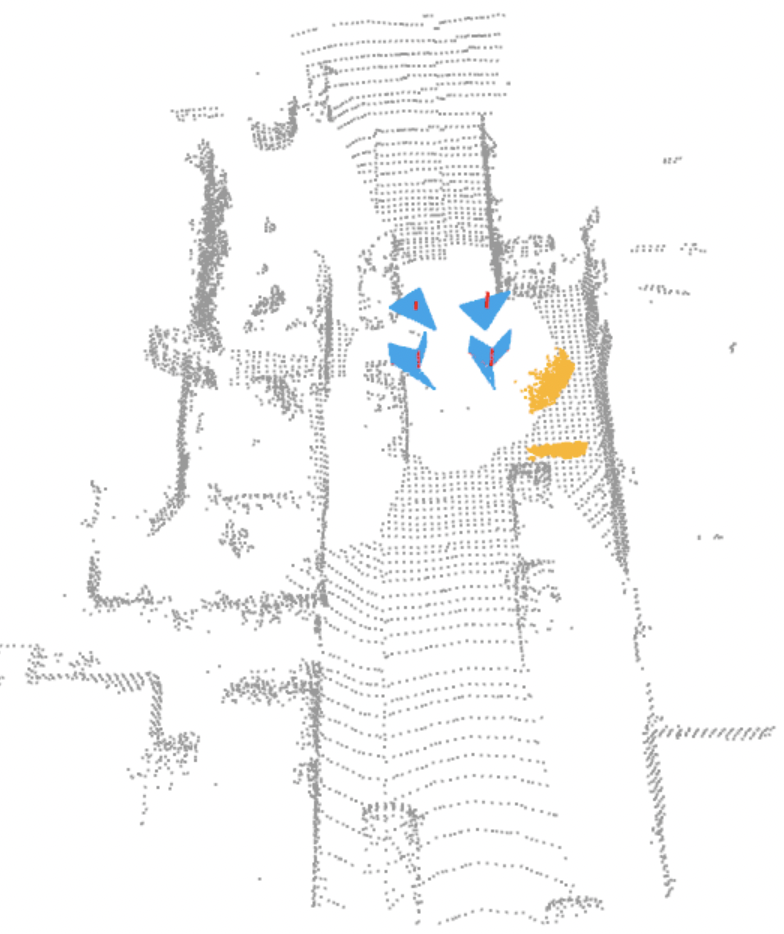}
        \label{fig:synthetic:scale_comparision}
    }
    \caption{Synthetic data generation steps. {We generate synthetic data through sampling primitive mesh models and augmenting real scan scattered points as noises.}}
    \label{fig:synthetic_data}
\end{figure}

\para{Synthetic Data Generation.}~\label{method:line:synthetic}
As discussed above in Sec.~\ref{sec:intro}, there are two types of reliable line segments to detect: 1) intersection between planes, and 2) poles. Hence, we choose to use the following two mesh primitives shown in Fig.~\ref{fig:synthetic_data:mesh} for simulating their local appearances, respectively. These two mesh models are first uniformly sampled into 4,000 points as Fig.~\ref{fig:synthetic_data:pc}, with 5\% relative 3-DOF positional perturbance added for each point. Then, to simulate possible background points nearby, we randomly cropped 40 basic primitives with each containing 1,000 points from real scans~\cite{geiger2012we}, and put them together to compose the final synthetic data. In total, we generated 5,000 synthetic point clouds with 5,000 points per each cloud.

% \begin{figure}
%     \centering
%     \includegraphics[width=0.8\textwidth]{./figure/synthetic_data.pdf}
%     \caption{Synthetic data generation}
%     \label{fig:synthetic_data}
% \end{figure}

% minipage中的figure无法单独加label

% There are two types of line segments we want to detect: poles and intersections between planes. Fig.\ref{fig:synthetic_data} shows the data generation process, including mesh generation, mesh sampling, and noise sampling. We connect the given vertices to build the mesh model, perform uniform sampling on the mesh surface to get the point cloud, and search the nearest point of the mesh edge to label the line feature. Since the distribution of the LiDAR points is complex and challenging to simulate, we sample some noise points from the LiDAR and add them to labeled point clouds.

\para{Scale-Invariant Line Segmentation.} We treat line detection as a point cloud segmentation problem, and the main challenge is the primitive scaling issue: In a real LiDAR frame, the density of points {decreases} with the scanning distance, and the voxel grid downsampling cannot fully {normalize the density} when the target feature is far away from the sensor. Moreover, our synthetic data generation also did not consider the scale of line segments {(as visualized in Fig.~\ref{fig:synthetic:scale_comparision} when put together). If such an issue is not handled,} the model will not produce reasonable predictions when the training and test data are on different scales.
% 
% 说一下如果不解决scaling issue会有什么问题。
% \ca{Discuss the bad influence of scaling issue.}

To address this issue, our network obtains scale invariance by eliminating the scale factor $s$ of the Sim(3) transformation and using relative distances, as:
\begin{equation}
\begin{aligned}
p' &= s \cdot \mathbf{R}p + t, \\
f  &= \frac{\sum_i (p' - p'_i)}{\sum_i \|p' - p'_i\|} = \frac{s \cdot \sum_i \mathbf{R}(p - p_i)}{s \cdot \sum_i \| p - p_i \|}.
\end{aligned}
\label{eq:rel_dist}
\end{equation}
In Eq.~\ref{eq:rel_dist}, we search $k = 20$ nearest points $\{ p_1, p_2, ..., p_k \}$ of a point $p$, and calculate {the scale-invariant local feature $f$ as} the ratio of the Manhattan distance to the Euclidean distance between $p$ and its neighbors. {The trade-off of such a feature definition is that} $f$ cannot reflect the position of the original point in the Euclidean space, so the transformation has information loss. Such an influence are further evaluated in Sec.~\ref{eval:line_seg}.
% . The scale-invariant method will be worse than the original method. We will discuss this issue in experiments (Sec.~\ref{eval:line_seg}).
% 说一下这样解决scaling issue有没有什么局限性？局限性的影响。
% \ca{Discuss the advantage and limitation of Eq.~\ref{eq:rel_dist}.}

\para{Model architecture.} We choose DGCNN~\cite{wang2019dynamic} {as our backbone, since it} directly encodes points and their nearest neighbors without complicated operations. Eq.~\ref{eq:raw_dgcnn} shows its local feature encoding function called $EdgeConv$~\cite{wang2019dynamic}, where $\textbf{x}_j$ is the $j$-th feature, $^{S}\textbf{x}_i$ is the neighbor of the $\textbf{x}_j$ in the feature space $S$, and $h$ is the learnable model.
\begin{equation}
    h \left(\textbf{x}_{j}, ^{S}\textbf{x}_{i}\right)=\bar{h}\left(\textbf{x}_{j}, ^{S}\textbf{x}_{i}-\textbf{x}_{j}\right).
    \label{eq:raw_dgcnn}
\end{equation}
%     h \left(\textbf{x}_{j}, \textbf{x}_{i}\right)=\bar{h}\left(\textbf{x}_{j}, \textbf{x}_{i}-\textbf{x}_{j}\right)
%
In the first $EdgeConv$ layer, $x$ represents the point coordinates in Euclidean space. In our implementation, we gather $k=20$ nearest neighbors of each points and calculate scale-invariant feature $f$. Then we turn the first $EdgeConv$ layer into:
\begin{equation}
    h \left({f}_{j}, ^{E}{f}_{i}\right)=\bar{h}\left({f}_{j}, ^{E}{f}_{i}-{f}_{j}\right).
    \label{eq:si_dgcnn}
\end{equation}
It replaces the coordinates in the Euclidean space with scale-invariant feature $f$, but $^{E}f_i$ is still the feature of $i$-th neighbor of point $p_j$ in Euclidean space, not the neighbor of $f_j$ in feature space. Since part of the information in the original Euclidean space has been lost when generating scale-invariant features, preserving the neighbor relationship in the original Euclidean space can reduce further information loss.

\begin{figure}
    \centering
    \includegraphics[width=0.88\textwidth]{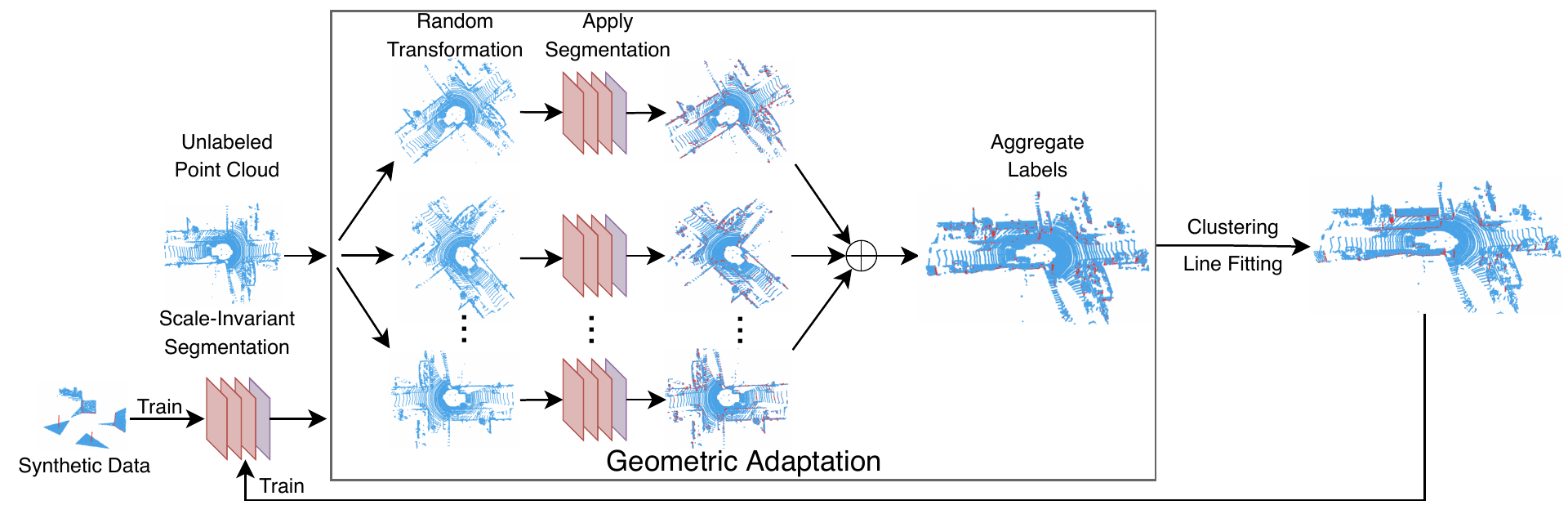}
    \caption{Automatic line labeling pipeline. We use geometric adaptation and line fitting to reduce the network prediction noise and improve model accuracy on real LiDAR scans through iterative training.}
    \label{fig:self_label}
\end{figure}

\para{Automatic Line Segment Labeling.}\label{method:line_label}
% We label LiDAR point clouds with a trained model on the synthetic dataset. Inspired by homography adaptation in SuperPoint~\cite{detone2018superpoint} and SOLD2~\cite{pautrat2021sold2}, we perform geometric adaptation on LiDAR point clouds. Specifically, we randomly rotate and translate the point clouds on the plane, predict their labels, and count the labels of each point. Points that have been detected more often are regarded as candidate points. Then we perform nearest-neighbor interpolation and line fitting to get the line feature label. The line segment detector is trained with the labeled LiDAR point clouds, and the geometric adaptation steps are repeated for three iterations to obtain more accurate line feature labels. \ca{Reorganize with method details.}
There is no available labeled line dataset of LiDAR scans, and performing manual labeling on point clouds is difficult. Hence, we build an automatic line labeling pipeline (Fig.~\ref{fig:self_label}). Inspired by homographic adaptation in SuperPoint~\cite{detone2018superpoint}, we perform geometric adaptation on LiDAR scans. First, we train a scale-invariant segmentation model {purely} on the synthetic data, and apply 2D transformations with a uniform distribution of $20m$ in XOY and $360^\circ$ in yaw to the LiDAR scans. Then, we use the trained model to predict labels on the perturbed data, aggregate the scan labels from all the perturbations and take the points that are predicted more than 80\% belonging to lines as candidate points. 
To cluster binary points into lines, we use the region-growth algorithm. The connectivity between points is defined through a $0.5m$ KD-Tree radius search. We use the labeled points as seeds, grow to nearby labeled points, and fit lines. 
% We take these steps iteratively until the line points remain unchanged.
% We perform region growing on the candidates to extract every possible line cluster. 
% After line fitting on the cluster, we iteratively gather the 5 closest points of each inlier, and again perform line fitting until the number of inliers remains unchanged.
{Once such line segments are extracted, we continue to refine} the segmentation model on the obtained labeled LiDAR scans. {We repeat the geometric adaptation 3 times to generate 12,989 automatically labeled LiDAR frames on the KITTI odometry sequences~\cite{geiger2012we}}.

\subsection{Joint Training of Line Segmentation and Description} \label{method:desc}

\para{Definition of Line Descriptors.} Different from the geometry definition which only requires two endpoints of a line segment. A descriptor for each line should convey local appearances through its all belonged points, since observed end points may be varied between frames due to possible occlusions.
%For the feature description, the point feature can use its feature vector as the descriptor, while the line feature is a collection of many points. It is difficult to learn the topology of the lines directly.
Therefore, we define the descriptor as an average of its all belonged points.

% \para{Problem Formulation.}
% There are many representations of line segments, including the two endpoints, two endpoints with a middle point, etc. Since the LiDAR point cloud has noise, and the features of some adjacent points are similar, using the endpoints to represent the line segment will introduce errors. We choose to extract all the points on the line segment. For the feature description, the point feature can use its feature vector as the descriptor, while the line feature is a collection of many points. It is difficult to learn the topology of the lines directly. Therefore, we use the line labels to calculate the average descriptor of the line as its descriptor.

\begin{figure}[t]
    \centering
    \includegraphics[width=0.72\textwidth]{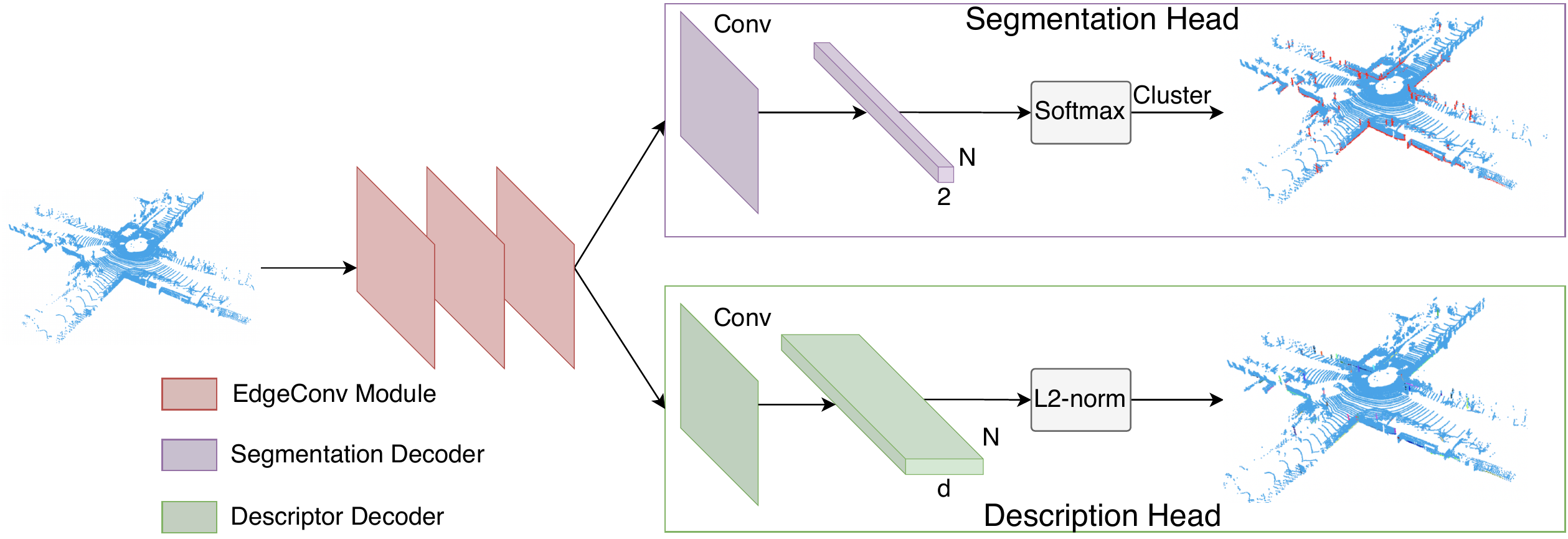}
    \caption{Network architecture. The network uses the EdgeConv~\cite{wang2019dynamic} module to extract features. The segmentation head and the description head predict the label and descriptor for each point, respectively.}
    \label{fig:network}
\end{figure}

\para{Network Architecture.}
Our network structure (Fig.~\ref{fig:network}) consists of a stacked three \emph{EdgeConv}~\cite{wang2019dynamic} layers for feature encoding, and two decoders for line segmentation and description, respectively.
Each $EdgeConv$ layer outputs a $N \times 64$ tensor used for 3-layer segmentation and description after a $MaxPooling$ layer. We use $ReLU$ for activation. 
The segmentation head turns the feature vector to a tensor sized $N \times 2$ after convolution ($N$ for the number of input points), and then obtains a {boolean} label per each point through a \emph{Softmax} layer, to predict whether it belongs to a line. The descriptor head outputs a tensor sized $N \times d$, and then performs \emph{L2-Norm} to get a $d$-dimensional descriptor. 

\para{Loss Functions.}
Our segmentation loss $\mathbf{L}_{seg}$ is a standard cross-entropy loss, and we follow~\cite{wang2019associatively} and ~\cite{bai2020d3feat} to build a discrimitive loss for the descriptor.
In detail, we first use the line segment label to get the mean descriptor $\mu$ of each line segment, and then use the $\mathbf{L}_{same}$ for each line to pull point descriptors towards $\mu$. The $\mathbf{L}_{diff}$ is proposed to make the descriptors of different lines repel each other. {In addition for a point cloud pair,} we calculate the matched loss $\mathbf{L}_{match}$ and the loss between the non-matched lines $\mathbf{L}_{mismatch}$. Each term can be written as follows:
\begin{equation}
\begin{aligned}
\mathbf{L}_{same}    &= \frac{1}{\mathbf{N}} \cdot \sum_i^{\mathbf{N}} \bigg( \frac{1}{|\mathbf{K}_i|} \cdot \sum_{j}^{\mathbf{K_i}} \left[\left\|\mu_{i}-d_{j}\right\|_{1}-\delta_{\mathrm{s}}\right]_{+}^{2} \bigg), \\
\mathbf{L}_{diff}    &= \frac{1}{|\mathbf{C}_\mathbf{N}^2|} \cdot \sum_{\left\langle i_a,i_b \right\rangle}^{\mathbf{C}_\mathbf{N}^2} \left[2 \delta_{\mathrm{d}}-\left\|\mu_{i_{A}}-\mu_{i_{B}}\right\|_{1}\right]_{+}^{2}, \\
\mathbf{L}_{match}   &= \frac{1}{\mathbf{N}} \cdot \sum_i^{\mathbf{N}} \left[\left\|\mu_{i}-\mu_{i}'\right\|_{1}-\delta_{\mathrm{s}}\right]_{+}^{2}, \\
\mathbf{L}_{mismatch}&= \frac{1}{|\mathbf{C}_\mathbf{N}^2|} \cdot \sum_{\left\langle i_a,i_b \right\rangle}^{\mathbf{C}_\mathbf{N}^2} \left[2 \delta_{\mathrm{d}}-\left\|\mu_{i_{A}}-\mu_{i_{B}}'\right\|_{1}\right]_{+}^{2},
\end{aligned}
\end{equation}
where $\mathbf{N}$ is the number of detected lines and $\mathbf{C}_\mathbf{N}^2$ stands for all pairs of two lines. $i$ and $j$ are two iterators, for lines and points on a line, respectively. $\mu_i$ is the aforementioned mean descriptor of a line, and $d_j$ is the descriptor of its related point descriptor $j$. $\mu_i'$ and $\mu_{i_B}'$ are mean descriptors in another associated point cloud, and $\delta_\mathrm{s}$ and $\delta_\mathrm{d}$ are the positive and negative margins. $[x]_+ = \max(0, x)$, and $\left\|\cdot\right\|_1$ for the L1-distance. Finally, we use $\omega = 2$ to balance the final loss $\mathbf{L}$ as:
\begin{equation}
    \begin{aligned}
    \label{eq_l_all}
        \mathbf{L} = \omega \cdot \mathbf{L}_{seg} + \mathbf{L}_{same} + \mathbf{L}_{diff} + \mathbf{L}_{match} + \mathbf{L}_{mismatch}.
    \end{aligned}
\end{equation}

\subsubsection{Line-based Registration}
Our network outputs labels and descriptors for each point. We first extract lines using steps in Section~\ref{method:line_label}. Then we perform descriptor matching to get line correspondences. The threshold of the matched descriptor is set to $0.1$. The transformation $\mathbf{T}$ for registering the source cloud $\mathbb{S}$ to the target cloud $\mathbb{T}$ is optimized by minimizing point-to-line distances of all line matching cost $\xi_i, i \in \mathbf{N}$: 
\begin{equation}
    \begin{aligned}
        \xi_i=\sum_j^{\mathbf{N}_i} \frac{\left|\left(\mathbf{T} \cdot p^\mathbb{S}_j - p^\mathbb{T}_{i_{e_0}}\right) \times \left(\mathbf{T} \cdot p^\mathbb{S}_j - p^\mathbb{T}_{i_{e_1}}\right)\right|}{\left| p^\mathbb{T}_{i_{e_0}} - p^\mathbb{T}_{i_{e_1}}\right|}
    \end{aligned}
    \label{point_to_line}
\end{equation}
where $p_j^\mathbb{S}$ is the line points in the source frame, $p_{i_{e_0}}^\mathbb{T}$ and $p_{i_{e_1}}^\mathbb{T}$ are endpoints of the matched line $\left\langle i_{e_0}, i_{e_1} \right\rangle$ of line $i$.

\section{Experiments}

\subsection{Network Training}

To begin with our generated synthetic data, we first train our line segmentation network using those synthetic point clouds with 50 epochs to converge. Then, to use {the auto labeling method} for generating sufficient and qualified real-world labeled scans, we obtain 12,989 LiDAR frames and iteratively train 100 epochs to refine these auto labeling results. Finally, we train our {whole line segmentation and description network} with 120 epochs to obtain the final applicable model for real-world scans.

We use scans including sequences 00-07 from the KITTI odometry dataset~\cite{geiger2012we}, with the last two sequences 06-07 for the validation set, and the rest 00-05 for the training set, to train our network. For each LiDAR frame, we voxelize the points cloud with $0.25m$ voxel size. We sample 20,000 points for evaluation and 15,000 points for training, since the kNN in $EdgeConv$ is $O(N^2)$ space complexity and consumes large memory in the training process. We calculate point-to-line distances following Eq.~\ref{point_to_line} on the line segments in Sec. \ref{method:line_label}. The line pair whose mean distance is within $0.2m$ will be selected as a line correspondence to calculate descriptor loss. We implement our network in Tensorflow~\cite{abadi2016tensorflow} with Adam~\cite{kingma2014adam} optimizer. The learning rate is set to 0.001 and decreases by 50\% for every 15 epochs. The whole network is trained on 8 NVIDIA RTX 3090 GPUs.

\subsection{Point Cloud Registration Test}

\para{Benchmarking.} We use sequences 08-10 from the KITTI odometry dataset~\cite{geiger2012we} to test the ability of our network on extracting line features and using them for point cloud registration. The preprocessing steps remain the same with our data preparation, and we choose to compare with traditional and learning-based methods for the global search registration. These traditional methods include ICP~\cite{besl1992method}, RANSAC~\cite{fischler1981random} and Fast Global Registration(FGR)~\cite{zhou2016fast}, are all implemented by Open3D~\cite{zhou2018open3d}. Specifically, The RANSAC and FGR use the FPFH~\cite{rusu2009fast} feature extracted from $0.25m$ voxel grid downsampled point clouds, and the max iteration is set to $4e^6$. Two learning-based methods include HRegNet~\cite{lu2021hregnet} and Deep Global Registration (DGR)~\cite{choy2020deep}, and they use ground-truth pose to calculate loss and predict the transformation directly through the network. PointDSC~\cite{bai2021pointdsc} learns to prune outlier correspondences. D3Feat~\cite{bai2020d3feat} and SpinNet~\cite{ao2021spinnet} extract salient features from point clouds. Our line-based registration extracts 18 line segments with 350 points per frame on average. For fair comparisons, the number of keypoints in learning-feature-based methods is also set to 350, while other parameters remain unchanged. 

\para{Metrics.} We use both the Relative Translation Error (RTE) and Relative Rotation Error (RRE)~\cite{lu2021hregnet} to measure the registration accuracy. Additionally, as a special reference for evaluating the success rate for global search registration methods, we treat those calculated transformations with relative error w.r.t. the ground truth smaller than $2m$ and $5^\circ$, as a successful attempt of registration.

% \linespread{1.2}
\setlength{\tabcolsep}{10pt}
\begin{table}[htbp]
\caption{Registration performance on KITTI dataset. {Our line segmentation and description method is highly competitive to the SOTA point-based approaches on the success rate, and both RTE and RRE can be refined with a subsequent coarse-to-fine ICP strategy.}}
\linespread{1.2}
    \begin{center}
    \begin{tabular}{l|cccc|c}
    \hline
    \multirow{2}{*}{} & \multicolumn{2}{c}{RTE (m)}              & \multicolumn{2}{c|}{RRE (deg)}            & \multirow{2}{*}{Recall} \\ \cline{2-5}
    & Mean           & \multicolumn{1}{c}{Std} & Mean           & \multicolumn{1}{c|}{Std} &                               \\ \hline
ICP~\cite{besl1992method}               & 0.417          & 0.462                    & 0.707          & 0.741                    & 11.30\%                        \\
FGR~\cite{zhou2016fast}               & 0.685          & 0.514                    & 1.080          & 0.921                    & 81.17\%                       \\
RANSAC~\cite{fischler1981random}            & 0.214          & 0.193                    & 0.924          & 0.907                    & 52.45\%                       \\ \hline
HRegNet~\cite{lu2021hregnet}           & 0.299          & 0.380                    & 0.712          & 0.643                    & 75.93\%                       \\
DGR~\cite{choy2020deep}               & 0.164          & 0.385                    & 0.226          & 0.569                    & 41.41\%                       \\
PointDSC~\cite{bai2021pointdsc}          & 0.187          & 0.225                    & 0.306          & 0.297                    & 44.98\%                       \\
SpinNet~\cite{ao2021spinnet}           & 0.183          & 0.142                    & 1.267          & 0.761                    & 93.98\%                       \\ 
D3Feat~\cite{bai2020d3feat}            & 0.088           & \textbf{0.043}           & \textbf{0.343} & \textbf{0.242}           & \textbf{98.90\%}              \\ \hline
SuperLine3D       & \textbf{0.087}  & 0.104                    & 0.591          & 0.444                    & 97.68\%                       \\ \hline
% SuperLine3D*       & \textbf{0.083}  & 0.088                    & 0.579          & 0.434                    & 98.28\%                       \\ \hline
% SuperLine3D*       & \textbf{0.082}  & 0.083                    & 0.578          & 0.427                    & 98.47\%                       \\ \hline
% SuperLine3D-th2*       & \textbf{0.083}  & 0.076                    & 0.582          & 0.432                    & 98.69\%                       \\ \hline
% SuperLine3D-th3*       & \textbf{0.085}  & 0.084                    & 0.584          & 0.432                    & 98.83\%                       \\ \hline
% SuperLine3D-th4*       & \textbf{0.085}  & 0.078                    & 0.584          & 0.430                    & 98.83\%                       \\ \hline
% SuperLine3D*       & \textbf{0.086}  & 0.079                    & 0.586          & 0.436                    & \textbf{98.92}\%                       \\ \hline

\end{tabular}
\end{center}
\label{tab:kitti_recall}
\end{table}
% 线方法平均每帧点云提18条线，350个点，  之前表里的d3feat spinnet都是1000个点的结果，现在换成了350点

\begin{figure}[htbp]
    \centering
    \includegraphics[width=\textwidth]{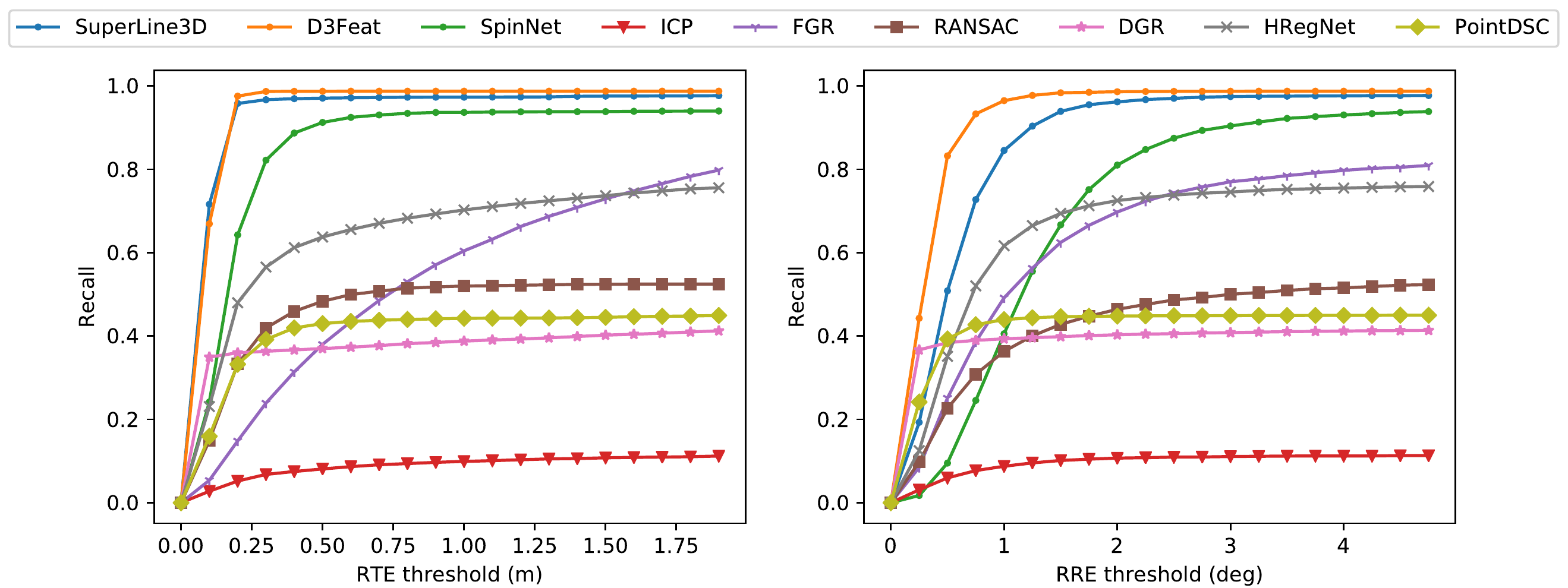}
    \caption{Registration recall with different RRE and RTE thresholds on the KITTI dataset. {The registration success rate of our line-based approach (blue) is close to the SOTA point-based approach D3Feat (orange) under different criteria.}}
    \label{fig:error_threshold}
\end{figure}

\begin{figure}[htbp]
    \centering
    \includegraphics[width=0.3\textwidth]{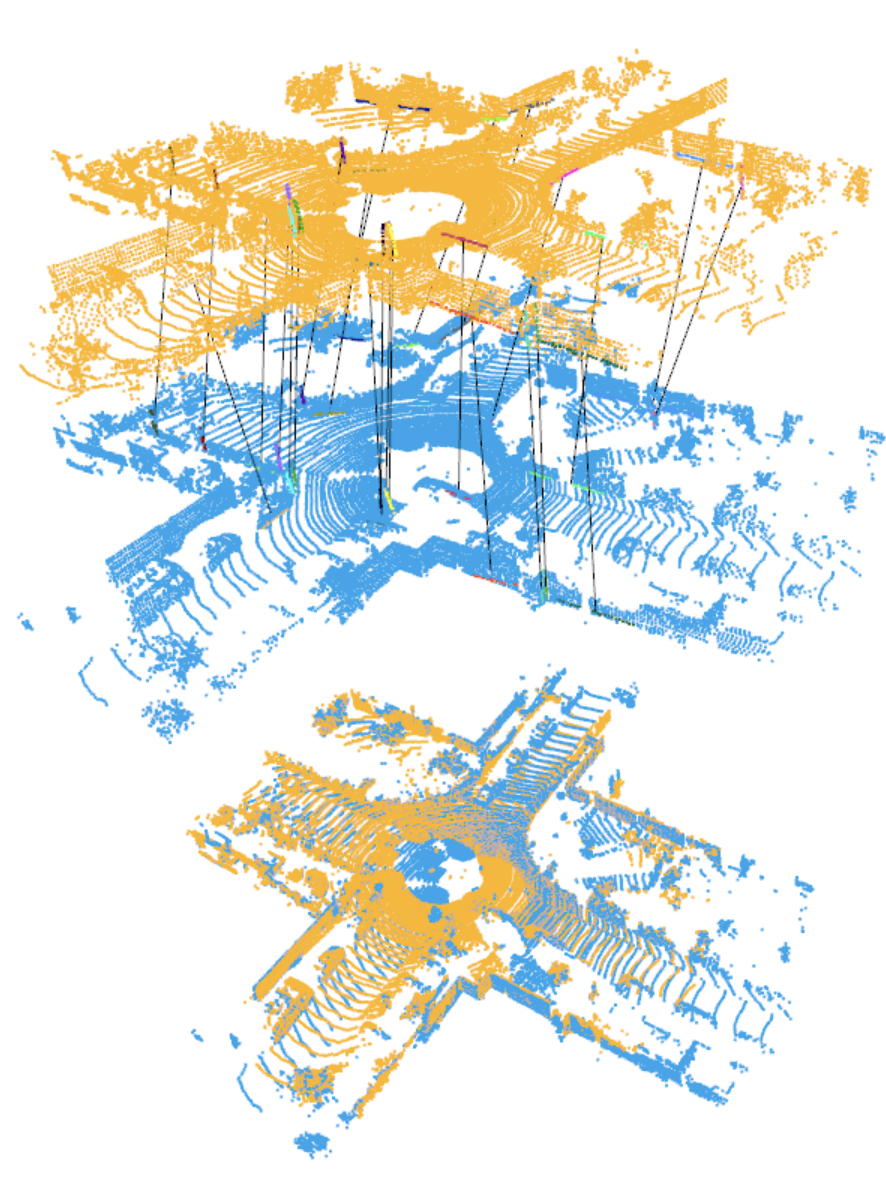}
    \includegraphics[width=0.35\textwidth]{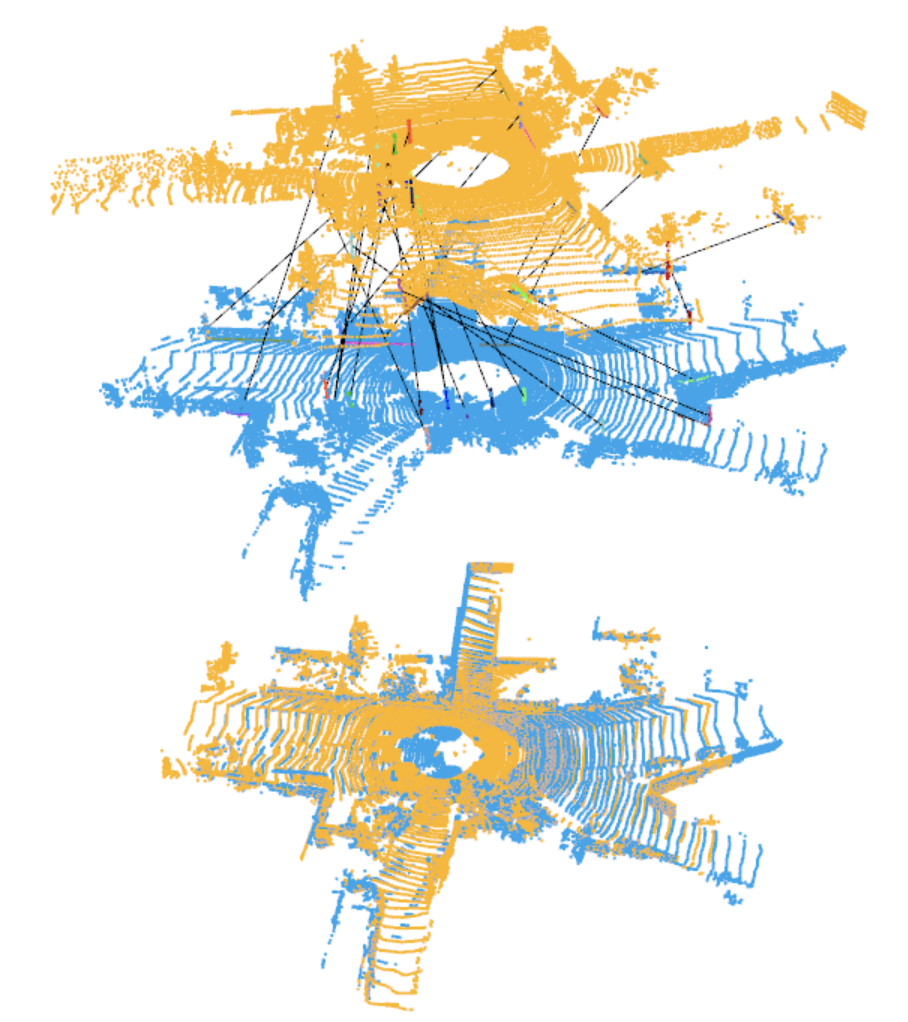}
    \includegraphics[width=0.33\textwidth]{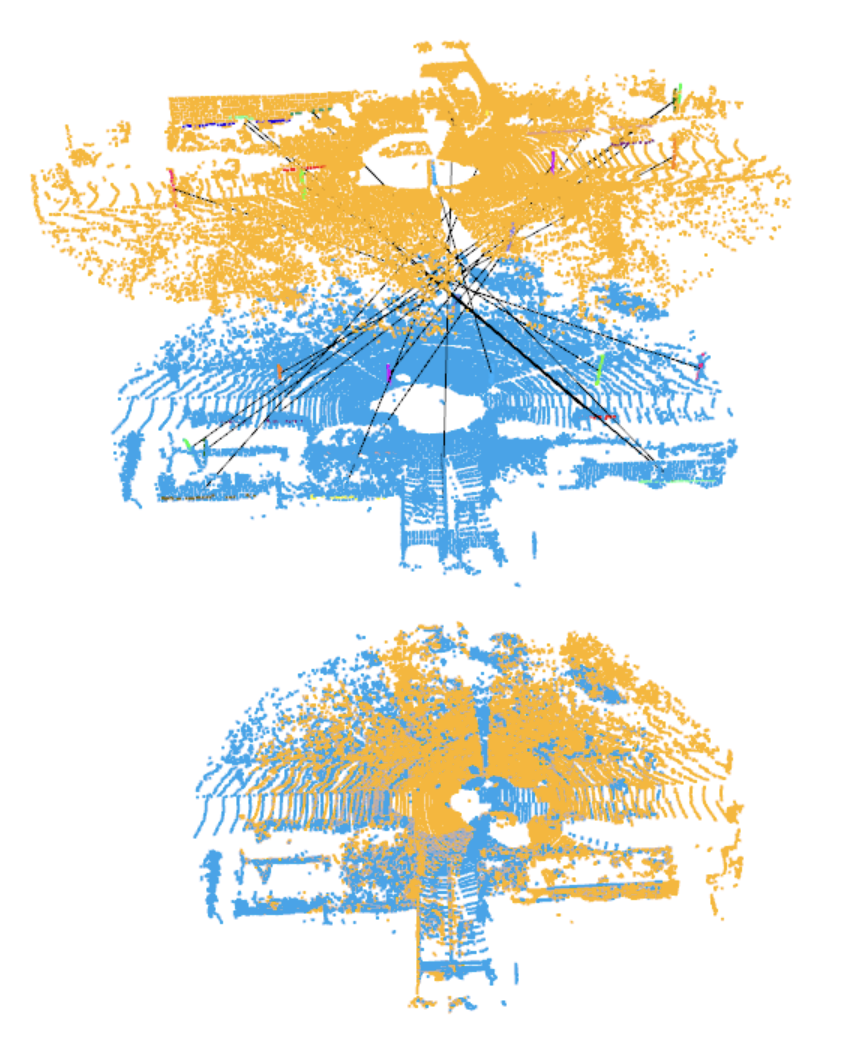}
    \caption{Qualitative visualization on KITTI test sequence. {Top: line associations between two LiDAR frames, Bottom: registration results of two frames.}}
    \label{fig:vis_kitti}
\end{figure}

\para{Results and Discussions.} Table~\ref{tab:kitti_recall} shows the registration performances. Under random rotation perturbation, the recall of ICP is only 11.3\%. The FGR and RANSAC methods based on FPFH features have higher recall but larger errors. The learning-based end-to-end methods HRegNet and DGR also drop in recall and accuracy when dealing with large perturbed scenarios. PointDSC relies on the feature model, and the features do not have full rotation invariance, so its performance also deteriorates. Fig.~\ref{fig:error_threshold} shows the registration recall with different error thresholds. SpinNet and D3Feat have better performances, with recall of over 90\%. Our line-based registration achieves comparable performance to point features, with a similar mean translation error and 1.22\% lower recall than D3Feat. Fig.~\ref{fig:vis_kitti} shows the visualization results on KITTI test sequence. Our method successfully registers point clouds under arbitrary rotation perturbations. We will give more results in supplementary materials.

\begin{figure}[htbp]
    \centering
    \includegraphics[width=\textwidth]{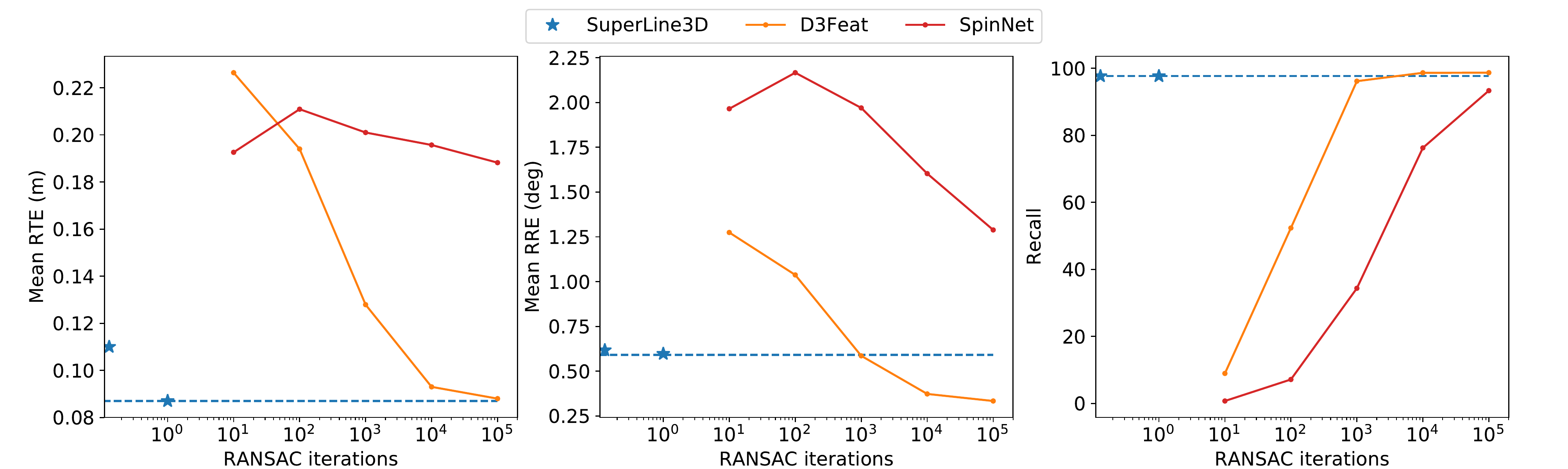}
    \caption{Registration performance with different RANSAC iterations. There are many mismatches in point feature correspondences, which leads to unstable results when the number of iterations is small.}
    \label{fig:ransac}
\end{figure}

% RANSAC is usually used to remove outliers and improve registration accuracy in the point feature-based registration method. As shown in Table~\ref{tab:ablation}, our line-based registration approach can achieve high accuracy without outlier removal. We use the estimated transformation to calculate the point-to-line distance of the matching lines in the two frames, directly remove the lines with an average distance greater than 1m, and re-estimate the transformation. Both rotation and translation accuracy are improved after outlier removal.
\para{Ablation on RANSAC iterations.} Point feature-based registration requires RANSAC to remove outliers and calculate the correct transformation. In the Table~\ref{tab:kitti_recall} and Fig.~\ref{fig:error_threshold}, the max iteration of RANSAC in the D3Feat and SpinNet operations are set to $5e^4$. {In contrast, our line-based registration does not rely on the RANSAC to filter erroneous matches: To perform outlier removal during transformation estimation, we calculate the line-to-line distances of line correspondences after the initial alignment, to remove the line correspondences with the mean distance greater than $1m$ and recalculate.} 

Fig.~\ref{fig:ransac} shows the performance of point cloud registration under different RANSAC iterations. The x-coordinates in the figure are logarithmic coordinates. Our method does not use RANSAC for outlier rejection, and we use a dashed line in blue as a reference when comparing with other methods requiring RANSAC post processing. The star near the y coordinates represents the original result, and the star with an x-coordinate of 1 is the result after outlier removal. Both D3Feat and SpinNet can not get accurate transformation without RANSAC until the max iteration exceeds 1,000.
% \para{Threshold of Success Rates.} 

\subsection{Line Segmentation Evaluation}\label{eval:line_seg}
To evaluate the scale-invariance of our base segmentation model, we train PointNet~\cite{qi2017pointnet}, PointNet++~\cite{qi2017pointnet++} and vanilla DGCNN~\cite{wang2019dynamic} on the synthetic dataset. The training set includes 4,000 synthetic point clouds normalized within $[0, 1]$. We test the trained model with point clouds scaled from 0.1 to 3.0. 

\begin{figure}[ht]
    \centering
    \includegraphics[width=0.8\textwidth]{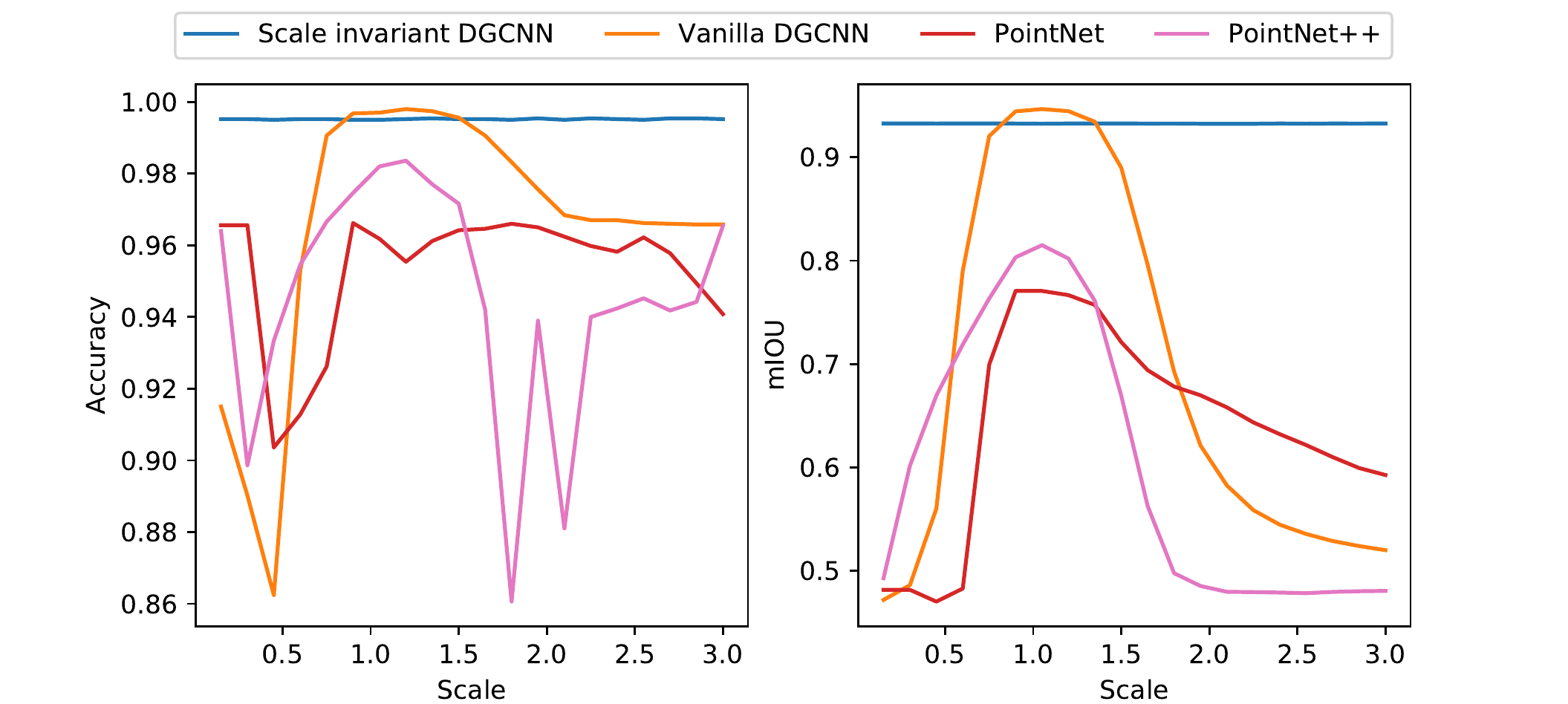}
    \caption{Accuracy and mIOU of network predictions under different scale disturbances. Our scale-invariant approach is stable under arbitrary scales, but is a little worse than the vanilla DGCNN in the original scale.}
    \label{fig:vis_si}
\end{figure}

\begin{figure}[ht]
    \centering
    \includegraphics[width=0.48\textwidth]{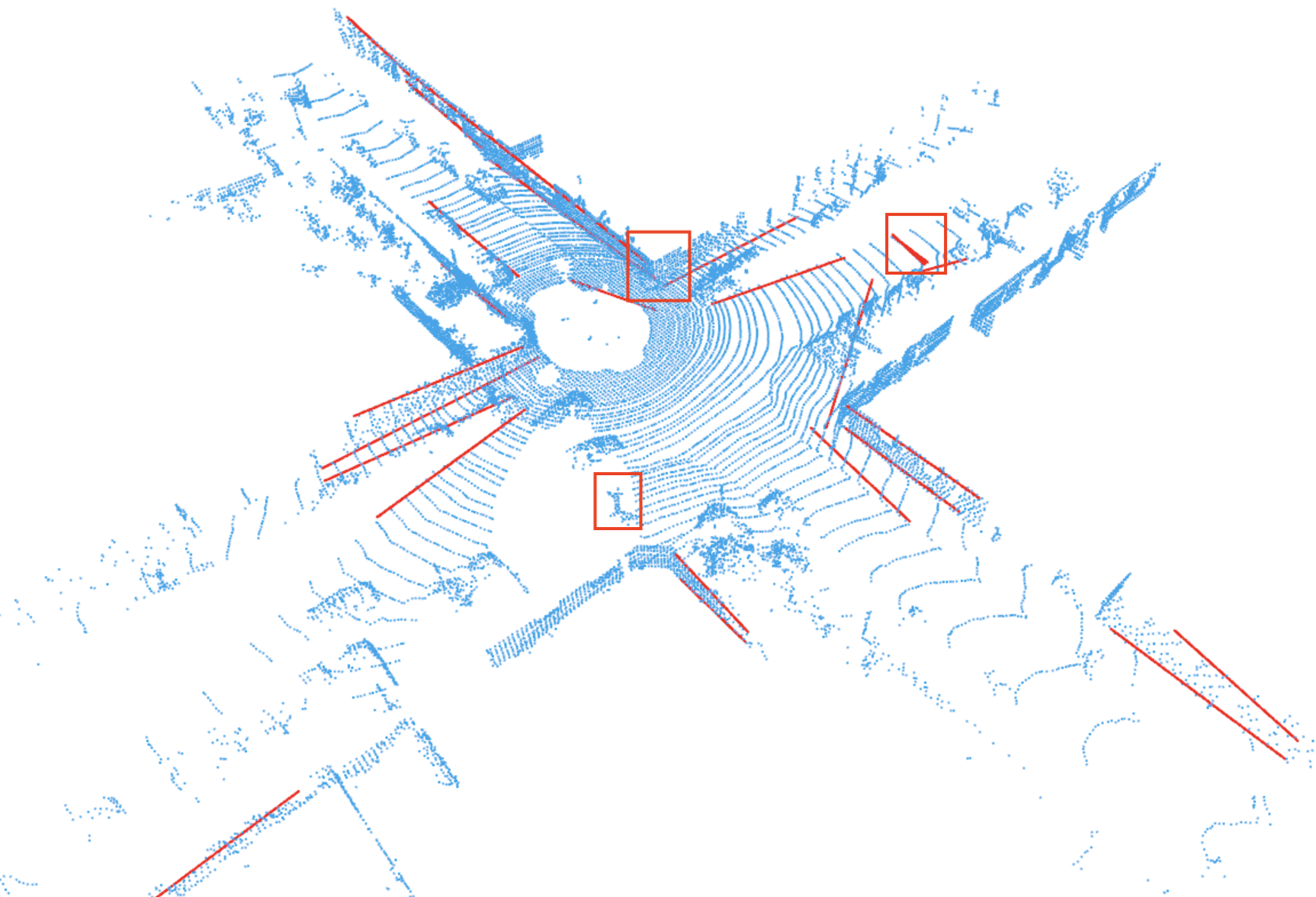}
    \includegraphics[width=0.50\textwidth]{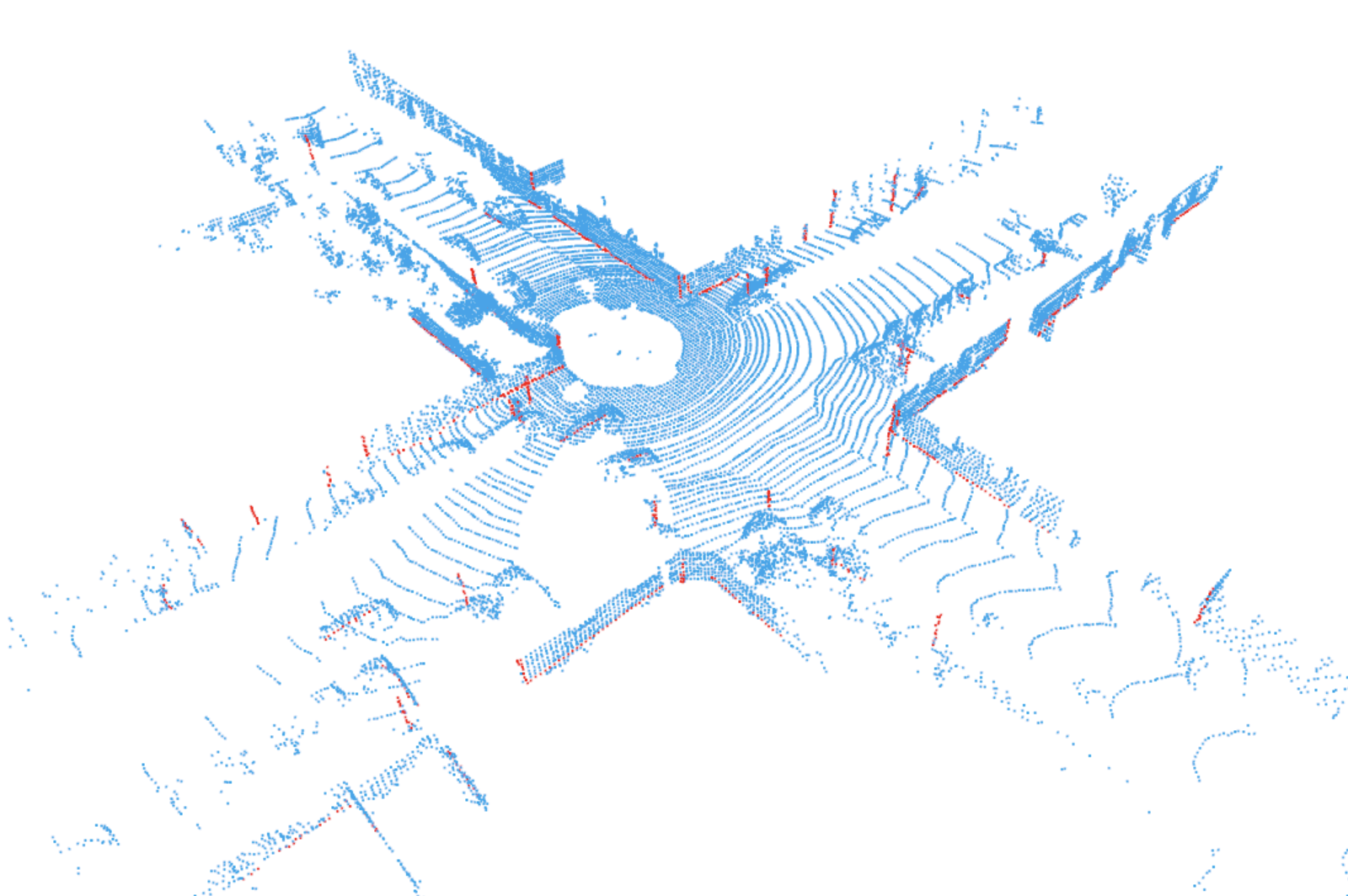}
    \caption{Qualitative visualization of line segmentation {between Lu et al.~\cite{lu2019fast} (left) and ours (right). Our method segments most of the poles and building edges.}}
    \label{fig:line_seg}
\end{figure}

Fig.~\ref{fig:vis_si} shows the accuracy and mIOU of network predictions. Methods without scale adaptation suffer from performance decrease when the scale changes. The vanilla DGCNN gets best accuracy and mIOU in small scale disturbance (0.8 to 1.6), while our scale-invariant approach is stable under arbitrary scales. We can find that when the scale is determined, using the scale-invariant approach will decrease the accuracy, so we only use it in synthetic data training. In the joint training of segmentation and description, we utilize the vanilla DGCNN instead.

Fig.~\ref{fig:line_seg} shows the qualitative visualization of our line segmentation compared with the only open-source 3D line detection method~\cite{lu2019fast} we found. Our method segments most of the lines, while the open-source one extracts LiDAR scan lines on the ground and cannot detect the poles.

\subsection{Generalization on Unseen Dataset}

To compare the generalization of learning feature-based models, we test our method against state-of-the-art point feature methods on the unseen Apollo Sourthbay dataset~\cite{L3NET_2019_CVPR} using the models trained on the KITTI dataset. We uniformly choose half of the point clouds from the SanJoseDownTown sequence as the source frames, select target frames every 5 frames, and add random yaw-axis rotation perturbances on the source frames. We get 8,296 point cloud pairs for evaluation. The data preprocessing of the point cloud is the same as the KITTI dataset.

\begin{figure}[htbp]
    \centering
    \includegraphics[width=0.35\textwidth]{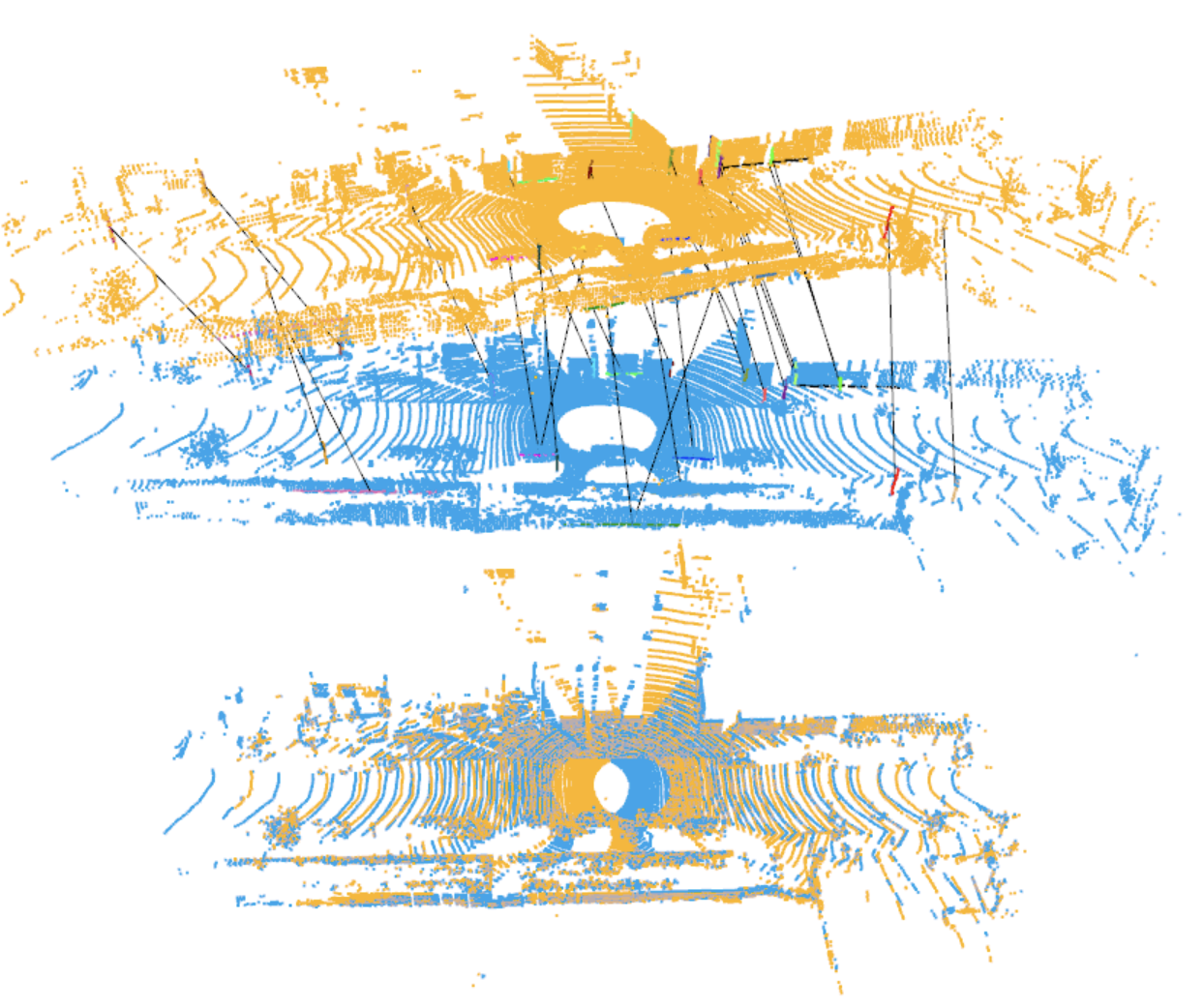}
    \includegraphics[width=0.25\textwidth]{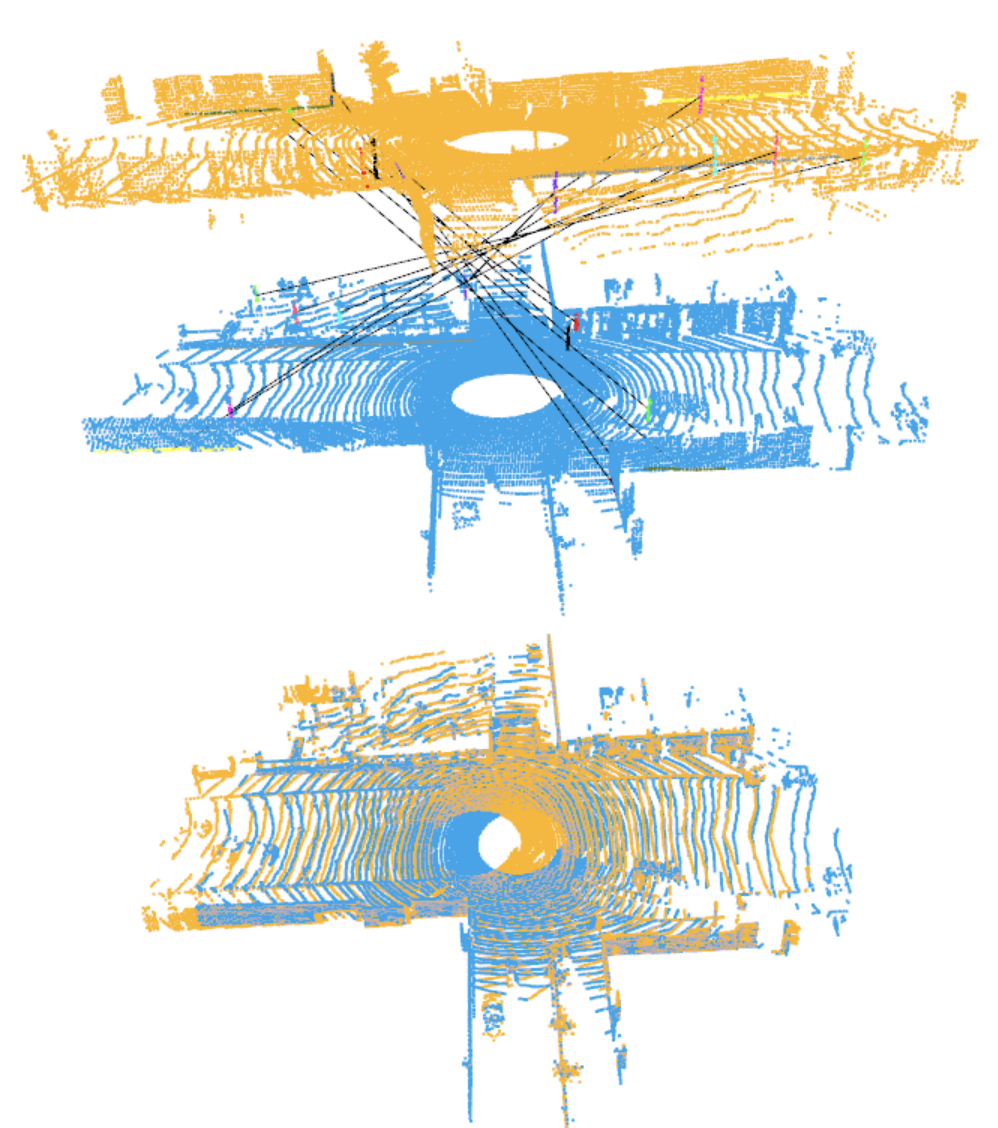}
    \includegraphics[width=0.33\textwidth]{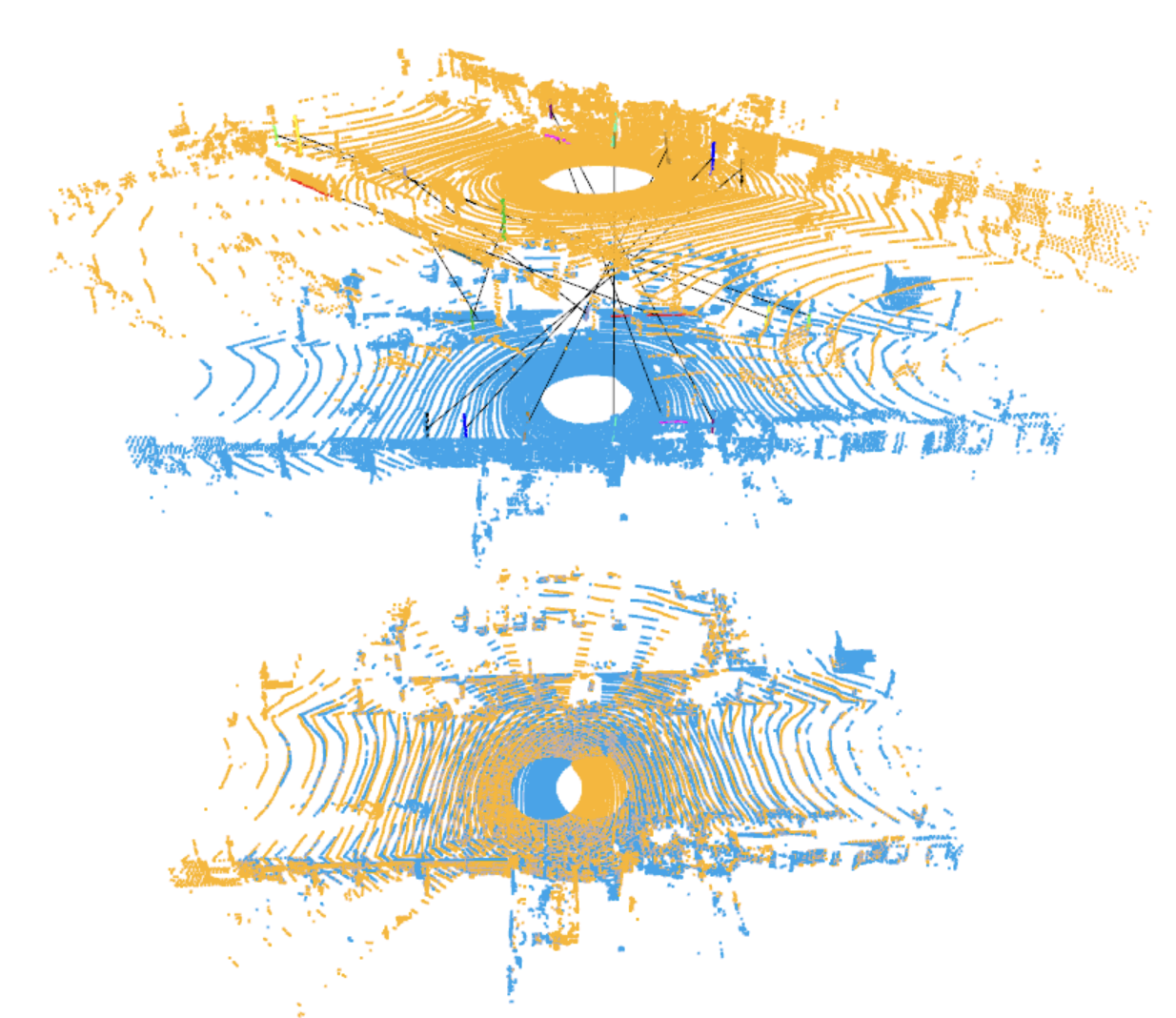}
    \caption{Qualitative visualization on Apollo SourthBay Dataset, SanJoseDowntown sequence. The majority of the line correspondences are stable poles, which helps reduce the translation error by a large margin.}
    \label{fig:vis_apollo}
\end{figure}

Table~\ref{tab:apollo} shows the point cloud registration results. On unseen datasets, all methods show a drop in recall. D3feat has the best performance, while the mean translation error of our method is the smallest one. Fig~\ref{fig:vis_apollo} shows qualitative visualization on the test data. There are more poles in this sequence, which is beneficial to our line-based registration.

\begin{table}[htbp]
    \centering
    \caption{Test on unseen Apollo SourthBay Dataset, SanJoseDowntown sequence.}
    \begin{tabular}{l|cccc|c}
    \hline
                & \multicolumn{2}{c}{RTE (m)} & \multicolumn{2}{c|}{RRE (deg)} & \multirow{2}{*}{Recall} \\ \cline{2-5}
                & Mean         & Std          & Mean           & Std           &                         \\ \hline
    SpinNet~\cite{ao2021spinnet}     & 0.199        & 0.203        & 1.207          & 0.874          & 75.66\%                        \\
    D3Feat~\cite{bai2020d3feat}      & 0.079        & \textbf{0.046}        & \textbf{0.206}          & \textbf{0.144}         & \textbf{95.94\%}                 \\
    % SuperLine3D & \textbf{0.040}    & 0.089        & 0.244          & 0.359         & 89.37\%                 \\ \hline
    SuperLine3D & \textbf{0.045}    & 0.107        & 0.262          & 0.402         & 93.84\%                 \\ \hline
    \end{tabular}
    \label{tab:apollo}
\end{table}
% 重复线特征较多，存在误匹配，可调

\subsection{Ablation Study}

\subsubsection{Skip Encoding}
The receptive field is directly related to the number of encoded features in the EdgeConv module. When $k$ is greater than 20, we can only set the batch size to 1 due to the enormous space complexity of EdgeConv. Its receptive field cannot be increased by increasing $k$. To this end, we utilize skip encoding. We gather $S \times k$ nearest neighbor features each time and select $k$ features with stride size $S$ for encoding. In this way, the receptive field increases $S$ times without consuming too much memory (gathering $S \times k$ nearest-neighbor features will also increase a little memory usage). In the experiments, we test the cases with stride 1 (nearest neighbor encoding), 2, 4, and 6. As shown in the Table~\ref{tab:ablation}, adjusting the stride to 4 reaches the best performance, since the local features cannot be well encoded when the stride is too large.

\subsubsection{Descriptor Dimension}

The descriptor dimension is one of the key factors for the feature matching performance, and the matching performance is poor when the dimension is low. Our network extracts dense descriptors. Each point has a descriptor of $d$ float numbers. It will take up a lot of storage space when its dimension is too large. Compared with the 16-dimension descriptor, the 32-dimension one has a more obvious improvement on the recall, while the 64-dimension descriptor has a small improvement. And increasing dimension to 128 only brings a smaller rotation error variance. Considering the average performances, we choose the 64-dimension implementation.

% Taking the 32-dimension descriptor as an example, the point cloud size of 20000 points is 240KB, and the descriptor size is 2.6MB, which is 11 times bigger than the point cloud. In the experiment, we compared the cases of 16, 32, and 64 dimensions. 

\begin{table}[ht]
    \centering
    \caption{Ablation study on stride and descriptor dimension.}
    \begin{tabular}{cc|cccc|c}
    \hline
                                                                                                         &     & \multicolumn{2}{c}{RTE (m)}     & \multicolumn{2}{c|}{RRE (m)}              & \multirow{2}{*}{Recall} \\ \cline{3-6}
                                                                                                         &     & Mean           & Std            & Mean           & \multicolumn{1}{c|}{Std} &                         \\ \hline
    \multicolumn{1}{l|}{\multirow{4}{*}{Stride}}                                                         & 1   & 0.092          & 0.134          & 0.594          & 0.449                    & 96.51\%                 \\
    \multicolumn{1}{l|}{}                                                                                & 2   & 0.088          & 0.116          & 0.595          & 0.465                    & 96.70\%                 \\
    \multicolumn{1}{l|}{}                                                                                & 4   & \textbf{0.087} & \textbf{0.104} & \textbf{0.591} & \textbf{0.444}           & \textbf{97.68\%}        \\
    \multicolumn{1}{l|}{}                                                                                & 6   & 0.134          & 0.216          & 0.783          & 0.757                    & 64.23\%                 \\ \hline
    \multicolumn{1}{l|}{\multirow{4}{*}{\begin{tabular}[c]{@{}l@{}}Descriptor\\ dimension\end{tabular}}} & 16  & 0.115          & 0.175          & 0.627          & 0.510                    & 87.56\%                 \\
    \multicolumn{1}{l|}{}                                                                                & 32  & 0.095          & 0.132          & 0.597          & 0.462                    & 95.28\%                 \\
    \multicolumn{1}{l|}{}                                                                                & 64  & \textbf{0.087} & \textbf{0.104} & \textbf{0.591} & 0.444                    & \textbf{97.68\%}        \\
    \multicolumn{1}{l|}{}                                                                                & 128 & 0.090          & 0.120          & 0.593          & \textbf{0.441}           & 96.70\%                 \\ \hline
    \end{tabular}
    \label{tab:ablation}
\end{table}

% \multirow{2}{*}{Outlier Removal}                                                         & w/o & 0.110        & 0.128        & 0.617 & 0.465                    & 97.68\%                 \\
%                                                                                 & w/  & 0.087        & 0.104        & 0.591 & 0.444                    & 97.68\%                 \\ \hline

\section{Conclusions}

This paper proposes the first learning-based 3D line feature segmentation and description method for LiDAR scans, which achieves highly-competitive performance to the point-feature-based methods in the point cloud registration. In the future, we will explore the usage of our deep learning line features on SLAM problems such as mapping, map compression, and relocalization. We will also optimize the network structure and reduce training resource consumption.

\subsubsection{Acknowledgments} This work is supported by the National Key R\&D Program of China (Grant No: 2018AAA0101503) and Alibaba-Zhejiang University Joint Institute of Frontier Technologies.

\clearpage
% ---- Bibliography ----
%
% BibTeX users should specify bibliography style 'splncs04'.
% References will then be sorted and formatted in the correct style.
%
\bibliographystyle{splncs04}
\bibliography{superline3d}
\end{document}